# Cluster-based Deep Ensemble Learning for Emotion Classification in Internet Memes

**Xiaoyu Guo[1]**

College of Economics and Management, Nanjing University of Aeronautics and Astronautics, Nanjing, 211100, China

**Jing Ma**

College of Economics and Management, Nanjing University of Aeronautics and Astronautics, Nanjing, 211100, China

**Arkaitz Zubiaga**

School of Electronic Engineering and Computer Science, Queen Mary University of London, London, E14 FZ, UK

## Abstract

Memes have gained popularity as a means to share visual ideas through the Internet and social media by mixing text, images and videos, often for humorous purposes. Research enabling automated analysis of memes has gained attention in recent years, including among others the task of classifying the emotion expressed in memes. In this paper, we propose a novel model, cluster-based deep ensemble learning (CDEL), for emotion classification in memes. CDEL is a hybrid model that leverages the benefits of a deep learning model in combination with a clustering algorithm, which enhances the model with additional information after clustering memes with similar facial features. We evaluate the performance of CDEL on a benchmark dataset for emotion classification, proving its effectiveness by outperforming a wide range of baseline models and achieving state-of-the-art performance. Further evaluation through ablated models demonstrates the effectiveness of the different components of CDEL.

## Keywords

Ensemble learning, Internet Memes, Emotion Classification, Clustering algorithm

## 1. Introduction

An internet meme is a form of viral information that spreads through social media and other online platforms, mixing various modes including text, image and video [1]. In the last few years, the growing ubiquity of interactive platforms such as Photoshop, Meitu, and Procreate have provided simple ways of creating internet memes, along with a number of online tools for automated generation of memes. Memes were one of the most typed words from January to March 2018 [3],

[1] **Corresponding author:**

Xiaoyu Guo, College of Economics and Management, Nanjing University of Aeronautics and Astronautics, Nanjing, 211100, China.

Email: xiaoyu.guo@nuaa.edu.cn

and a growing number of people search for memes on the Google search engine [1]. A meme is generally based on an image template, with a text on top of it. For example, the Boromir memes with the caption of "one does not simply...", shown in Figure 1, are derived from the movie Lord of the Rings. Indeed, the same background template can lead to the production of several memes with very different meanings by simply altering the texts.

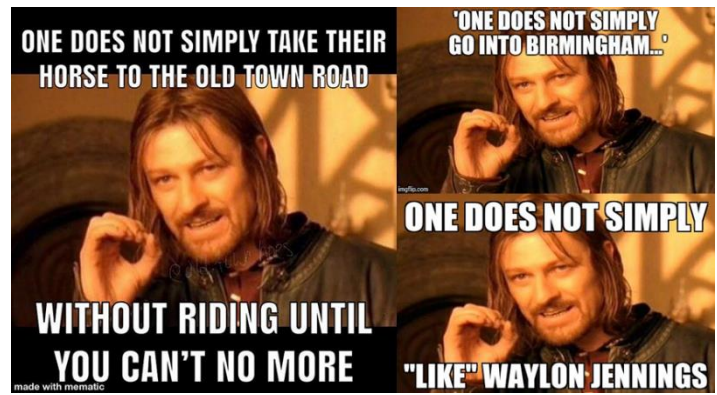

**Figure 1.** Examples of memes from the Memotion dataset using the popular Boromir format.

A body of recent research has looked into automated approaches for analyzing internet memes [4]. One of the popular tasks associated with internet memes is the classification of emotion expressed in the memes [8], which is the focus of our research. Classification of emotion can be very useful for many other tasks. For example, the attitudes toward a breaking news story can be determined by using an aggregation of features, including emotion. Where social media has become an important source for research into emotions, consideration of memes becomes important as not all the emotions expressed in social media are text-based. In this work we tackle the emotion classification task for the static image-with-caption class of memes, which is the most common type of meme in social media [11]. Despite their popularity, automated analysis of memes is still in its infancy due to the challenge of exploiting multi-modal models that make effective use of both image and text [9].

The existing research regards the meme as the multi-modal public opinion and use general multi-modal models to analyze memes. Research proposing ad-hoc methods to deal with memes, exploiting its inherent multi-modal features, is much more limited. In tackling emotion classification for memes, we take the distinctive features of memes into account and propose to build on the fact that some memes can share common features with other memes, such as having a similar background or similar facial expressions. To make the most of this inherent characteristic of memes, we propose a novel clustering-based deep ensemble learning (CDEL) method. To build CDEL, we integrate a deep learning model, a clustering algorithm, and a classification algorithm. The intuition of incorporating a clustering algorithm is that it will group the memes by facial features, which can enhance the model by exploiting the emotions expressed in faces. The extraction of facial expressions can enable its application to new, unseen memes.

The implementation of the hybrid CDEL model involves three key challenges that we address: (1) given that image-with-caption memes are composed of pictorial and textual components, **choosing the features to be extracted from each component** for clustering is important for the model performance; (2) **selecting clustering and classification algorithms** that lead to an optimal joint performance, for which in our case we make use of a "local optimization" approach; and (3)

**combining the clustering algorithm and the deep learning model**, which can have a big impact on model performance.

The CDEL model achieves state-of-the-art performance on a well-known benchmark dataset released as part of the SemEval-2020 task 8 [9]. Through an ablation study, we also demonstrate the effectiveness of the different components of CDEL, which all together lead to optimal performance. Our work highlights the importance of leveraging an ensemble model to process multi-modal objects like memes.

The rest of this paper is organized as follows: In Section 2 we discuss meme classification and its related work. Then in Section 3, we propose the CDEL method and give the details of the implementation. We then show the experimental process and results on evaluating the performance of CDEL and present their implications in Section 4. Finally, we conclude the paper in section 5.

## 2. Related work

We discuss related work in two different areas: internet meme classification and ensembles based on clustering models.

### *2.1. Internet meme classification*

Research on internet meme classification is at an early stage. Next, we discuss two main directions of work related to image-based internet meme classification: (1) detecting whether an image is a meme or not, and (2) internet meme emotion analysis. While our research focuses on the latter, we briefly give an overview of the former due to the relevance of the approaches used.

#### *2.1.1. Meme detection*
The meme detection task consists in detecting whether an image posted in social media constitutes a meme or is something else. This research generally assumes that images input to the model contain some form of text on them. One common practice to extract the text from these images is to use Optical Character Recognition (OCR) [13]. However, the majority of OCR approaches are trained from images containing black text on a white background, which doesn't directly apply to memes which have more complex image backgrounds and different text colors.

Research has tried to overcome this challenge by performing OCR on meme-like images. For example, Beskow et al. [11] proposed to preprocess the images by (1) converting the image to grayscale, (2) binarizing the image, and (3) inverting every bit in the binary image array. This is effective to then use tools such as Tesseract [14] to extract the text. On top of that, they proposed a multi-modal deep learning model, combining features extracted from texts, images and faces, with the ultimate goal of classifying images as memes vs. non-memes. Alternative methods to meme detection include that by Zannettou et al. [15], who use perpetual hashing and clustering techniques, and the one by Dubey et al. [16], who proposed an algorithm based on sparse representations and deep learning to produce a rich semantic embedding for image-based memes.

In our case focused on meme emotion classification, the dataset consists of only memes and not other kinds of images, where the textual content of the memes is already provided, hence our work doesn't directly focus on the prior steps of detecting whether an image is a meme and on the OCR.

*2.1.2. Meme emotion analysis*

We discuss research in internet meme emotion analysis in three tasks: (1) emotion classification, (2) identifying the type of emotion expressed, and (3) detecting hateful memes.

The goal of **emotion classification** is to classify internet memes into one of positive, negative, or neutral. Priyashree et al. [17] used K-Nearest Neighbours (KNN) and Multinomial Naïve Bayes (MNB) for emotion classification, whereas Amalia et al. [18] used OCR Tesseract in combination with a Naive Bayes classifier [14]. The latter achieved a competitive accuracy of 75%, however lacking sufficient data for detailed analysis of results. Wu et al. [10] further used slang and sentiment lexica for the emotion classification task.

The Memotion Analysis [9] was a shared task that took place as part of SemEval 2020. In this paper we make use of the dataset released as part of its Task A, which consists in classifying memes into one of positive, negative, or neutral. To consider the achievements of the shared task participants in our experiments, we also compare the top-ranked submissions to the task as baselines in this paper. The shared task attracted a total of 583 participants. Most of the participants used multi-modal deep learning methods combining image and textual features, while few of them based their models on traditional machine learning models and on uni-modal approaches considering only text or image. Keswani et al. [19] achieved the top position in the shared task by using a Feed-Forward Neural Network (FFNN) with Word2vec embeddings that relied solely on textual features, ignoring the image; this method outperformed other models using combinations of both text and image, as well as methods leveraging only image. In this paper we demonstrate that an effective combination of both image and text can lead to improved performance over the sole use of text. We provide further detail of the top performing models in the task in Section 'Baseline models'.

The task of **identifying the type of emotion expressed in a meme** consists in classifying an input meme as humorous, sarcastic, motivational or offensive. Costa et al. [20] proposed a Maximum Entropy classifier for recognizing humorous memes. Their model achieved high performance for the negative class (F1 = 99.9%), with a substantially lower performance on the positive class (F1 = 63.7%). To encourage more researchers to put effort into emotion analysis of memes, a shared task was organised as part of SemEval (Task 8), which proposed to perform the aforementioned 4-class classification [9].

The goal of **detecting hateful memes** consists in classifying memes as hateful vs. not hateful. Sabat et al. [8] use Bidirectional Encoder Representations from Transformers (BERT) and the deep convolutional network with 16 layers proposed by Visual Geometry Group (VGG-16) to process texts and images of memes, respectively. To boost the performance of hate speech detection in memes, Kiela et al. [21] organised the Hateful Memes Challenge. In addition, in a study characterizing internet memes, Beskow et al. [11] found that the nature of memes enables easier avoidance of censorship, in turn enabling increased virality within and across social media platforms compared to other kinds of images [22].

In summary, existing research on internet meme classification has considered a wide range of traditional machine learning and more contemporary deep learning models. However, to our knowledge, their combination has not been studied to date, not least in enriching deep learning models with the outputs from clustering models. In studying this combination in the context of memes, our study introduces a first-of-its-kind model that attempts to leverage the best of both worlds.

## 2.2. Ensemble models leveraging clustering

The majority of research focuses on single deep learning methods such as BERT [23], XLNet [24], and Densenet [25] for classification. In contrast to these, ensemble learning has succeeded in various tasks [26]. The idea behind ensemble learning is to process separate elements (e.g. modes) with different algorithms, subsequently fusing them to produce the final, joint decision. Since image memes are composed of pictorial and textual components, we propose to use ensemble learning as an effective means for making the most of both modes.

Clustering algorithms aim to learn groups of common features shared by different data instances in an unsupervised manner [27]. For example, to extract features from internet memes, Dang et al. [29] used a clustering algorithm to group textual memes into a few clusters, helping with the detection of misinformation and their spread.

Over the last few years, researchers used clustering algorithms to improve the performance of classification based on machine learning [30], rarely using deep learning to combine clustering and classification approaches. Our model CDEL combines the output of a clustering algorithm with a deep learning model to make the most of both components, which is to the best of our knowledge the first model of this kind for meme classification.

## 3. Methodology

### 3.1. The CDEL framework

We depict the CDEL framework in Figure 2.

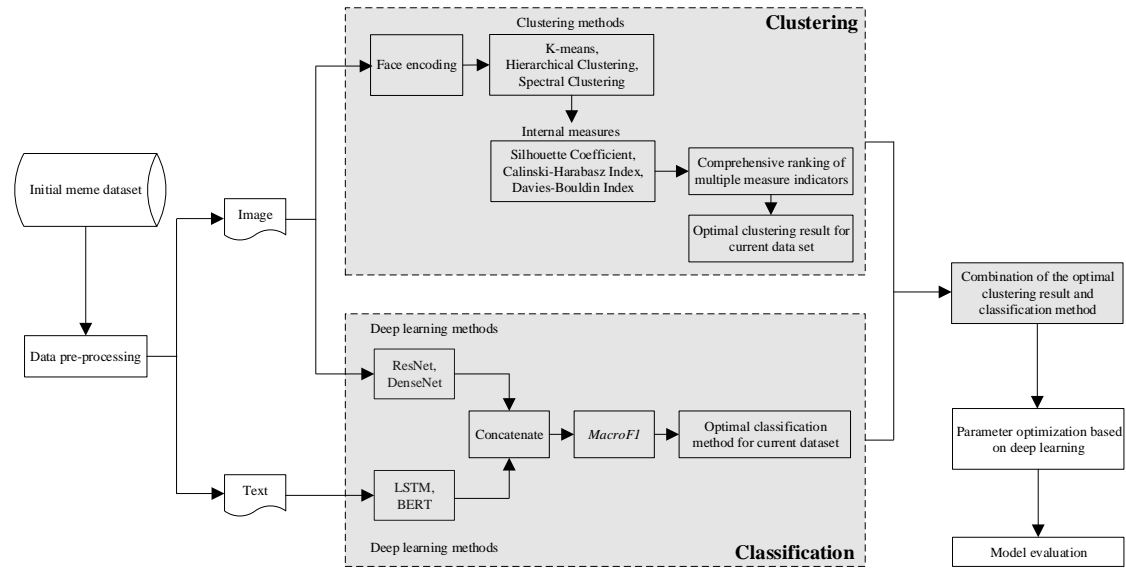

**Figure 2.** Architecture of the CDEL framework.

First, we select the optimal clustering algorithm and deep learning classification models in parallel. We evaluate different clustering algorithms and classification models to empirically determine the optimal ones. Subsequently, we concatenate the clustering outcome with the features extracted from the deep learning model.

## 3.2. Cluster-based emotion classification model

In what follows, more detail on the clustering and classification components of the framework is described.

Given that faces are a vital element of memes, we first extract face encodings. The reason why we do not use color features is that most of the memes have dark backgrounds so it is hard to extract distinctive color features. Facial features are extracted by face recognition, an open-source face detection library, proposed by Adam Geitgey and available on GitHub [35]. These facial features are then input to the clustering algorithm, which enables determining clusters based on facial expressions. Images that don't contain a face in them are not processed by the clustering algorithm and they are consequently put into a new cluster.

We tried three of the most common image clustering algorithms [36]: K-means, Hierarchical Clustering, and Spectral Clustering. K-means and Hierarchical Clustering are built using the SciPy library [37] and Spectral Clustering using the scikit-learn library [38]. The clustering yields, for each meme, the ID of the cluster it belongs to.

In parallel to the clustering process, we perform multimodal classification of the memes. We tested four different joint Deep Neural Network (DNN) classifiers: (1) Residual Network (ResNet) [39] and Long Short Term Memory (LSTM) [40] (2) ResNet and BERT [23] (3) Densely Connected Network (DenseNet) [25] and LSTM, and (4) DenseNet and BERT.

Figure 3 shows the joint classification architecture, in this case taking ResNet and LSTM (#1 above) as an example. We use the pre-trained model ResNet to extract image features and remove the last dense fully connected layer. In parallel, LSTM takes word embeddings and a hidden vector as the input and yields a new hidden vector. Next, the new hidden vector and ResNet features are concatenated into a feature vector. Finally, the concatenated layer is used as the input to a fully connected layer, followed by the activation of *sigmoid* (or *softmax*) for generating the probability of the image for pertaining to a class. The *sigmoid* and *softmax* functions are defined as:

$$\sigma(t) = \frac{1}{1+e^{-t}} \quad (1)$$

$$softmax(z_i) = \frac{e^{z_i}}{\sum_{m=1}^{M} e^{z_m}} \quad (2)$$

where $z_i$ is the output of node $i$ and $M$ denotes the number of classes.

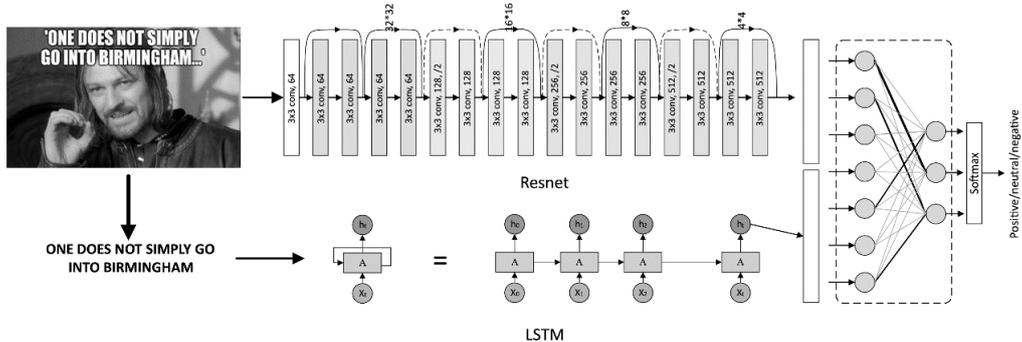

**Figure 3.** Joint deep learning model for meme classification.

While optimising the combination of clustering and classification models, we dealt with two additional challenges: (1) selection of the number of clusters, and (2) choosing the optimal combination of algorithms. In what follows, we describe how we addressed these two challenges.

## *3.3. Selection of the optimal number of clusters*

To determine the number of clusters $c$, we propose a multi-evaluation method based on Hierarchical Clustering. There is no need to determine the number of clusters when using Hierarchical Clustering. However, this algorithm has another vital parameter $t$, which is the threshold that helps forming flat clusters. Hence, the first step is to find the optimal $t$. The reason why we select $t$ rather than $c$ to optimize is that the range of $t$ values is much smaller than those of $c$ for a large number of memes.

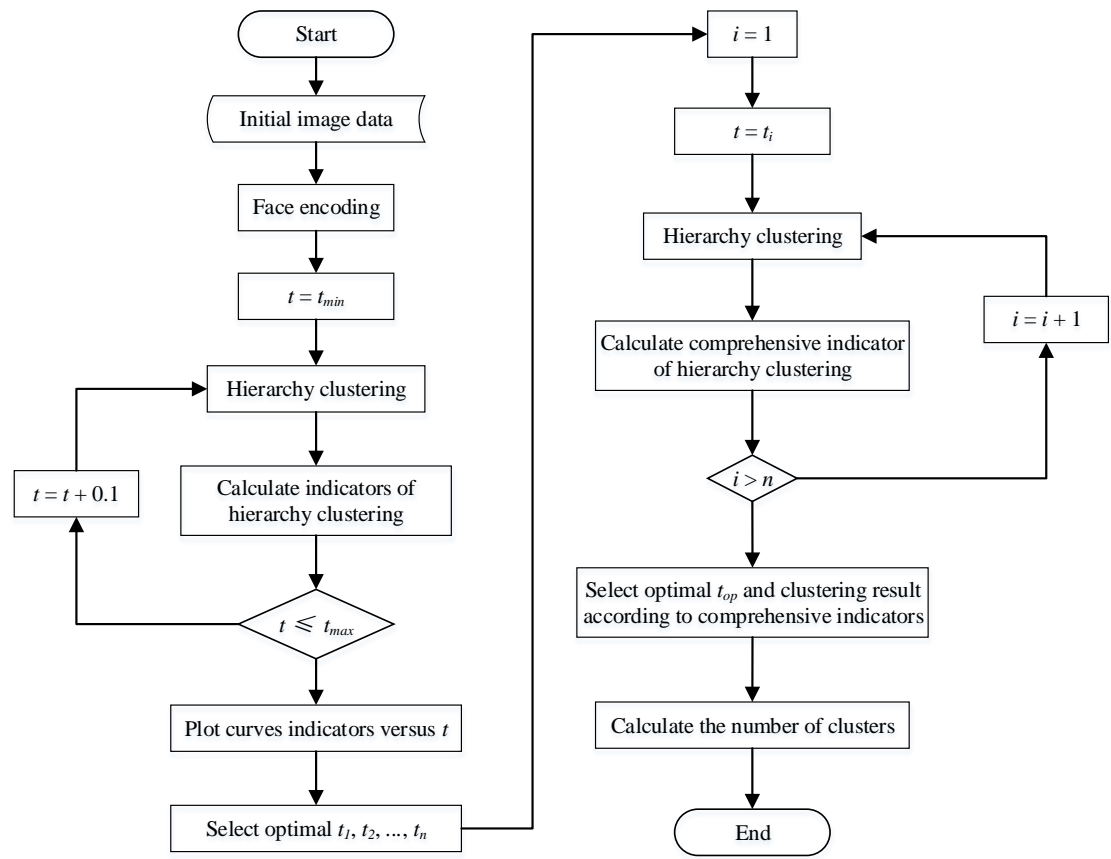

**Figure 4.** Selection of the optimal number of clusters.

Figure 4 shows the diagram we follow for determining $c$, which includes the following steps:
1. After extracting face encoding features from initial image data, we populate a matrix with pairwise distances between images. The range of possible values of $t$ is then determined by the minimum ($t_{min}$) and maximum ($t_{max}$) values in the matrix.
2. We test all candidate values of $t$ (with a step size of 0.1) into the Hierarchical Clustering algorithm, producing the evaluation indicators (see Section 'Evaluation measures' for details on the indicators used).
3. For each evaluation indicator, we draw the curve indicator against the value of $t$, ultimately selecting the optimal $t$ based on the Elbow method.

4. Now, we have the optimal $t_1, t_2, ..., t_n$, where $n$ denotes the number of indicators. For each optimal $t$, we pass it to the Hierarchical Clustering algorithm and calculate the comprehensive indicator.
5. We then select the optimal $t_{op}$ based on the comprehensive indicator.
6. Based on the clustering output produced by using $t_{op}$, the optimal number of clusters $c$ is inferred.

## *3.4. Combination strategy*

In this section, we propose a novel strategy to combine a deep learning model and a clustering algorithm (see Figure 5 and Figure 6). For each meme, we concatenate text features, image features extracted by a pre-trained deep learning model, and one-hot encoded cluster ID produced by the clustering algorithm (see Algorithm 1 for a flow chart of CDEL).

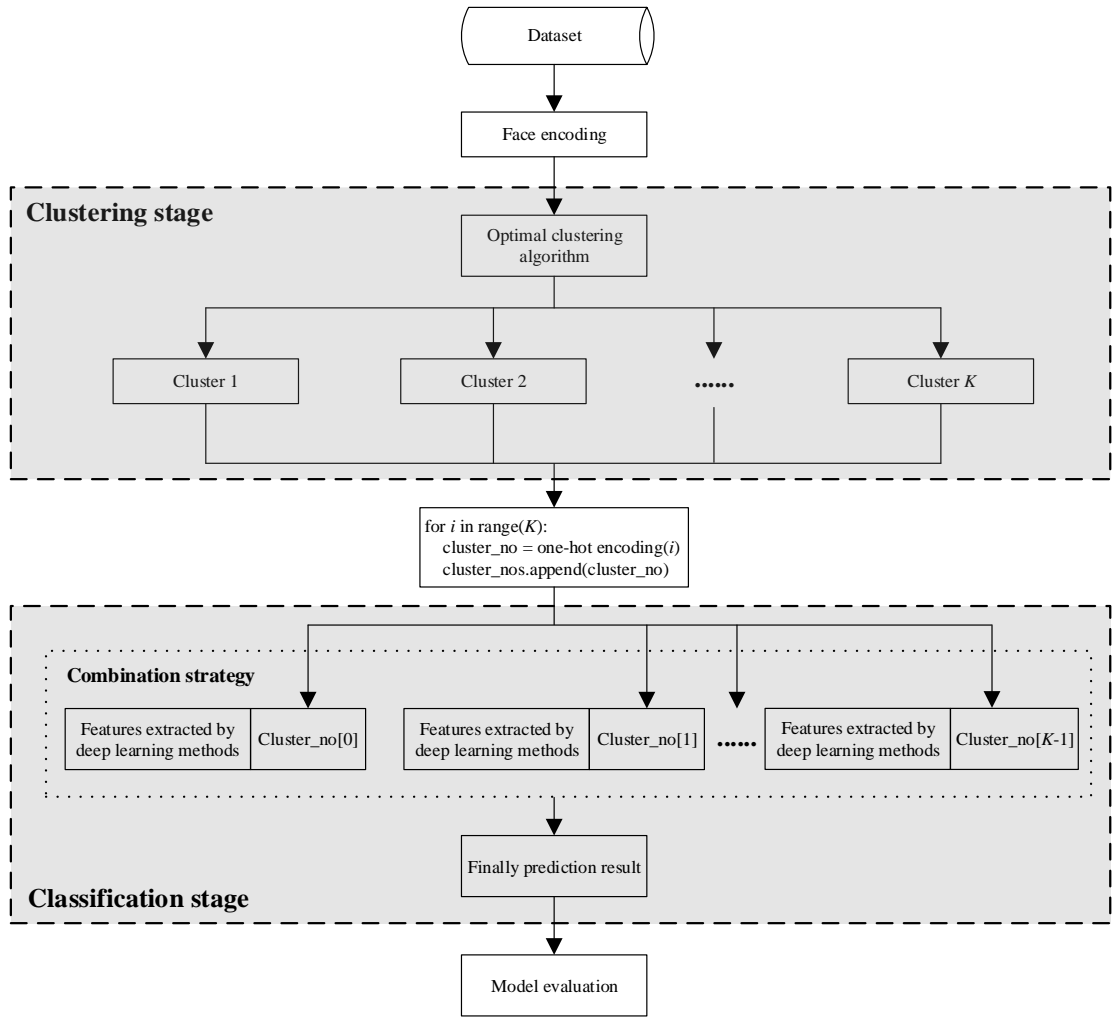

**Figure 5.** A simplified diagram of the combination strategy of clustering and deep learning methods.

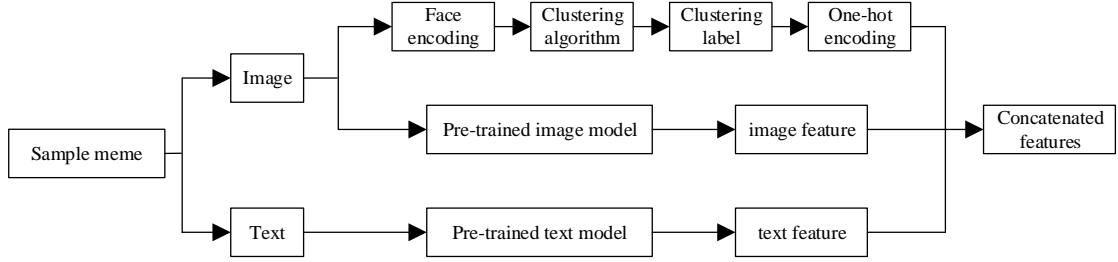

**Figure 6.** A simplified diagram of the combination of three components.

| **Algorithm 1. CDEL Algorithm** |
|---|
| Input: Training set $D_{train}$; testing set $D_{test}$ |
| Output: the labels of the samples in $D_{test}$ |
| Steps: |
| 1. Traverse the candidate clustering algorithm set to select the optimal clustering algorithm and parameters suitable for the current dataset $D_{train}$; |
| 2. Traverse the candidate multi-modal deep learning model set to select the optimal classification model and hyperparameters suitable for the current dataset $D_{train}$; |
| 3. Use the optimal clustering algorithm to divide the training set $D_{train}$ into $c$ clusters, get the clustering labels of $D_{train}$, and get the clustering label set $\{N_1, N_2, ..., N_c\}$; |
| 4. **For** $i$=1 to $c$ do<br>    Generate one-hot encoding $O_i$ for $N_i$<br>   **End for** |
| 5. **For** $i$=1 to $len(D_{train})$ do<br>    Generate text and image embedding $e_i$ using optimal classification model for $D_{train}[i]$;<br>    Concatenate $e_i$ with corresponding clustering label's one-hot encoding $O_i$;<br>   **End for** |
| 6. Concatenate embeddings on a dense fully connected layer, followed by a *sigmoid* (or *softmax*) activation, and train the model; |
| 7. Prediction of sample memes in $D_{test}$ using trained model and evaluation of sample labels in $D_{test}$. |

## 4. Experiments

### *4.1. Dataset*

We use the meme emotion classification dataset from the Memotion Analysis competition [9]. The organisers of the competition collected these images through querying the Google search engine, using 52 different keywords such as Hillary, Trump, etc.. The emotion class labels were annotated through crowdsourcing by Amazon Mechanical Turk workers. We focus on task A of the competition, which consists in establishing the emotion of memes as one of positive, negative or neutral. Each entry in the dataset contains the following relevant fields: image, text, and sentiment.

The dataset contains a total of 8,870 samples, distributed into 5,271 positive (59.4%), 2,795 neutral (31.5%) and 804 negative (9.1%) instances. Following Guo et al. [2], where the text field of a meme is null in the dataset, we populate it with the text extracted from the image.

For the experimentation, we rely on the training and test datasets as split by the organizers. In addition, we extract a stratified sample from the training set, which becomes the development set (Dev). Table 1 shows the distribution of labels across training, dev and test sets.

|          | Train | Dev   | Test  | Sum   |
|----------|-------|-------|-------|-------|
| Positive | 3,089 | 1,071 | 1,111 | 5,271 |
| Neutral  | 1,634 | 567   | 594   | 2,795 |
| Negative | 469   | 162   | 173   | 804   |
| Sum      | 5,192 | 1,800 | 1,878 | 8,870 |

Table 1. Dataset statistics.

## *4.2. Parameter setting*

All the hyper-parameters in the model were adjusted by the performance on the training set. The batch size is set to 128, the learning rate is set to $2e^{-5}$ and Adam is used as the optimizer. The Dropout rate is 0.3. For the LSTM model, we set the number of layers as 1 and the number of units as 128. For the BERT model, we choose the uncased BERT-base model, including an encoder with 12 self-attention heads, 12 Transformer blocks, and a hidden size of 768. Since the dataset is imbalanced, we use the strategy of Logit Adjustment [41] to overcome this problem. This strategy is implemented by changing the loss function and can be directly used in Keras[2]. Therefore, the loss function we use in our experiments is sparse_categorical_crossentropy_with_prior.

## *4.3. Baseline models*

To assess the performance of our CDEL model against other competitive approaches, we also report the results of a number of baseline models that achieved top performance at the shared task and can be deemed state-of-the-art models based on the assessment on the Memotion dataset:

- **N-gram+VGG-16+SVM** [42] use n-grams as text features and VGG-16 for image features, concatenate the two kinds of features and classify the final features using Support Vector Machine (SVM).
- **ALBERT+VGG-16+multi-task** [43] combines A Lite BERT (ALBERT) and VGG-16 into a unified architecture. They also use multi-task learning to combine all of the Memotion tasks.
- **LSTM+VGG-16**[44] extract text and image features using LSTM and VGG-16 respectively, concatenate them and apply a feed-forward neural network to produce the final label.
- **Inception-resnetV2** [45] use inception-resnetV2 to extract only image features and obtain the final classification result.
- **BERT + VGG using Softmax** [46] use BERT to process text and VGG-16 to obtain the image features, and fuse the two kinds of features using two locally-connected layers, with 512 and 256 hidden units respectively, following a *softmax* activation.

---
[2] https://kexue.fm/archives/7615

- **KNN** [47] perform various experiments and the results show that considering either of the text or image is better than combination of both. Therefore, their optimal model uses a KNN on image embeddings.
- **Guoym** [48] use two ensemble types as their combination strategy. On one hand, they perform a data-based ensemble, with splitting the training data into five subsets. Each of these subsets is then used to train a model, which leads to five models which vary in terms of data and parameters. These five models are then combined through an ensemble. On the other hand, for the feature-based ensemble, they build five different models: textual models with Bidirectional Gated Recurrent Unit (Bi-GRU), Embedding from Language Models (ELMo) and BERT, image features with Resnet, and the combined text-vision model. Features extracted by these models are combined into the feature-based ensemble. Both ensembles are then combined together for the final model.
- **FFNN** [19] perform a Feed-Forward Neural Network (FFNN) approach with Word2vec embeddings, whose features are only based on text.

## *4.4. Evaluation measures*

### *4.4.1. Performance metrics for clustering*

The methods to assess the performance of clustering include two categories: external indicators and internal indicators [49]. Given that external indicators require ground truth cluster labels, which are not available in our datasets, we use internal indicators.

Internal indicators rely on inherent features of the dataset to evaluate the clustering, e.g. average similarity within clusters, average similarity between clusters or overall similarity. The most common internal indicators are Silhouette Coefficient (SC), Calinski Harabasz Score (CHS), and Davies Bouldin Index (DBI).

To assess the overall performance of the clustering algorithm, we define a Comprehensive Indicator (CI). Taking into account that SC and CHS yield positive scores but DBI yields a negative score, we define CI as:

$$CI = SC_N + CHS_N - DBI_N \qquad (3)$$

where $SC_N$, $CH_N$ and $DBI_N$ are the normalization format of *SC*, *CHS* and *DBI*. The normalization method is

$$x(k)_N = \frac{x(k) - min(X)}{max(X) - min(X)} \qquad (4)$$

### *4.4.2. Performance metrics for emotion classification*

Following the official evaluation proposed in the Memotion shared task, we adopt the *MacroF*1 score. To compute this score, we first calculate the *F*1 score for each class, as follows:

$$F1 = \frac{2TP}{2TP + FP + FN} \qquad (5)$$

where *TP* is the number of correct positive samples, *FP* is the number of negative samples predicted as positive, and *FN* is the number of positive samples predicted as negative.

For the final *MacroF*1 score, the average of the *F*1 scores for all classes is computed:

$$MacroF1 = \frac{F1_{positive} + F1_{negative} + F1_{neutral}}{3} \qquad (6)$$

## 5. Results

### 5.1. Selecting the optimal number of clusters

The range of candidate values of $t$ we get is (0.1, 1.1), which are fed to test the Hierarchical Clustering. Figure 7 shows the curves of internal indicators (SC, CHS, and DBI) against $t$. According to each of the indicators, we select: (1) $t_1$ as 0.8 according to SC since the SC is a positive score, (2) given that for CHS we cannot find the optimal t from the middle of the curve, we choose the optimal $t_2$ as 1.1, and (3) given the DBI yields a negative score, $t_3$ is set to 0.7 based on the Elbow method.

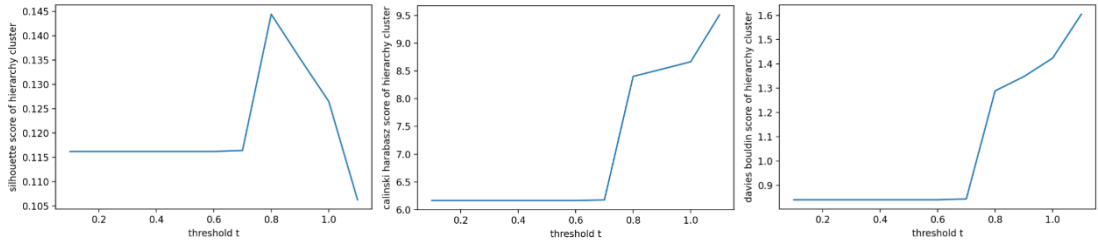

**Figure 7.** Curves of SC (left), CHS (middle), and DBI (right) versus $t$.

Table 2 shows the final evaluation results by using the selected set of $t$ candidates, i.e. {$t_1$ = 0.8, $t_2$ = 1.1, $t_3$ = 0.7}. Based on these results, we select 0.8 as the best value of $t$ and consequently 1,562 as the best value of $c$ according to CI.

|  | SC | CHS | DBI | CI | $c$ |
|---|---|---|---|---|---|
| $t_1$=0.8 | 0.140 | 7.858 | 1.304 | **1.011** | **1,562** |
| $t_2$=1.1 | 0.106 | 9.107 | 1.601 | 0 | 2,567 |
| $t_3$=0.7 | 0.110 | 5.869 | 0.852 | 0.118 | 2,561 |

**Table 2.** The value of SC, CHS, DBI, CI, and corresponding $c$ with the optimal $t$ set on Memotion Analysis Dataset. The bold value of CI is the best among the three experiments and the bold 1,562 is the best value of $c$ according to CI.

### 5.2. Selecting the optimal clustering algorithm

Table 3 summarizes the experimental results of the image clustering algorithms with $c$ = 1,562, showing an overall best performance of Hierarchical Clustering among three clustering algorithms.

|  | SC | CHS | DBI | CI |
|---|---|---|---|---|
| K-means | 0.070 | 6.651 | 1.312 | 0.331 |
| Hierarchical Clustering | 0.140 | 7.858 | 1.304 | **1.078** |
| Spectrum Clustering | -0.013 | 2.147 | 1.210 | 0 |

**Table 3.** Comparison of different clustering algorithms. The bold value of CI is the best among the three clustering algorithms.

## 5.3. Selecting the optimal deep learning classification model

After determining the optimal clustering algorithm, we determine the optimal joint pre-trained deep learning model through experiments on the development set. We report the experimental results in Table 4, as *MacroF*1 scores. From these results, we conclude with DenseNet+LSTM as the most accurate of the models under study.

| Model | *MacroF1* |
|---|---|
| ResNet+LSTM | 0.3240 |
| ResNet+BERT | 0.3441 |
| **DenseNet+LSTM** | **0.3444** |
| DenseNet+BERT | 0.3374 |

**Table 4.** *MacroF1* for different joint pre-trained deep learning models on Memotion Analysis development Dataset. The bold 0.3444 is the best value of *MacroF1* and the bold DenseNet + LSTM is the best model name according to the value of *MacroF1*.

## 5.4. Results of experiments with CDEL and comparison with existing methods

We use the optimal combination of clustering and classification models determined through the above process. In the results shown in Table 5, we see that CDEL outperforms all the other models, with a relative improvement of 3.4% (or absolute improvement of 1.22%) over the best baseline, FFNN. This sets our model as the state-of-the-art model for this task and dataset, which we analyze in more detail through ablated models and a breakdown of the results in what follows.

| Model | *MacroF1* |
|---|---|
| N-gram+VGG-16+SVM [42] | 0.3400 |
| ALBERT+VGG-16+multi-task [43] | 0.3454 |
| LSTM+VGG-16[44] | 0.3460 |
| Inception-resnetV2[45] | 0.3469 |
| BERT + VGG using Softmax[46] | 0.3475 |
| KNN[47] | 0.3500 |
| Guoym[48] | 0.3520 |
| FFNN[19] | 0.3547 |
| **CDEL** | **0.3669** |

**Table 5.** *MacroF1* scores for CDEL and baseline models. The bold 0.3669 indicates that CDEL is the best model.

Table 6 shows the results of the ablation study, where we compare a range of combinations that only use one of the modes (image or text), as well as combinations using CDEL's clustering component or not. Looking at the text-only models, we see that there is a minor performance difference between LSTM and BERT. After adding the Cluster algorithm, the performance difference between LSTM+Cluster and BERT+Cluster for the CDEL models using text only increases. When we look at image-only models, however, DenseNet obtains a substantial improvement over ResNet. Among the CDEL models using image only, DenseNet achieves a further improvement over ResNet, hence proving DenseNet as a more competitive approach.

|  | Model | *MacroF1* |
|---|---|---|
| Text-only | LSTM | 0.3248 |
|  | BERT | 0.3203 |
| Image-only | DenseNet | 0.3432 |
|  | ResNet | 0.3289 |
| Text-only CDEL | LSTM+Cluster | 0.3415 |
|  | BERT+Cluster | 0.3331 |
| Image-only CDEL | **DenseNet+Cluster** | **0.3587** |
|  | ResNet+Cluster | 0.3328 |

**Table 6.** Ablation study. The bold 0.3587 indicates that DenseNet+Cluster is the best model.

The addition of the clustering component to text-based models (i.e. text-only CDEL over text-only) leads to improvement, and likewise image-based models (i.e. image-only CDEL over image-only). Still, none of the ablated models achieves the performance of the complete CDEL model incorporating both modes and the clustering component, achieving an overall best 0.3669.

In order to ensure the robustness of the output results, we split all the data into five parts and perform a five-fold cross-validation. The results are 35.90%, 36.03%, 36.00%, 36.39% and 35.89%, separately. The final result is 36.04% by taking the arithmetic mean.

These results demonstrate the potential of the CDEL model and the need to incorporate all of its components for optimal performance. In this case, CDEL with all of its components achieved a score of 0.3669, outperforming all of the ablated variants. This is because CDEL adds the outcome of the Hierarchical Cluster algorithm as an additional feature to the Densenet+LSTM model, boosting the MacroF1 from 0.3444 (Densenet+LSTM) to 0.3669 (CDEL), which highlights the important benefits of incorporating the clustering model.

## *5.5. Breakdown of CDEL results*

Delving into the predictions of the CDEL model, we focus on the confusion matrix shown in Table 7. The test set contains 1,878 instances, among which 889 are correctly classified by CDEL, leading to an accuracy of 47.34%. Still, we observe some common mistakes of the model in distinguishing between classes, particularly between the positive and neutral classes, with several actually positive instances predicted as neutral (310) and actually neutral instances predicted as positive (287).

|  |  | Predicted class | | |
|---|---|---|---|---|
|  |  | Negative | Neutral | Positive |
| Actual class | Negative | 27 | 49 | 97 |
|  | Neutral | 82 | 225 | 287 |
|  | Positive | 164 | 310 | 637 |

**Table 7.** Confusion matrix of the proposed CDEL on Memotion Analysis test Dataset.

Further looking into the class-specific results of CDEL, we show the Precision, Recall and F1 scores for each class in Table 8. We observe that the overall best performance is obtained for the positive class, whereas the negative class gets the lowest performance. One of the main factors that likely has a big impact on this is that the positive class is the most frequent (59.4% of the test set) and the negative class is the least frequent (9.1%).

|  | Precision | Recall | F1 |
| --- | --- | --- | --- |
| Negative | 0.0989 | 0.1561 | 0.1211 |
| Neutral | 0.3853 | 0.3788 | 0.3820 |
| Positive | 0.6239 | 0.5734 | 0.5976 |

**Table 8.** Precision, Recall and F1 of each class.

## *5.6. Error Analysis*

To identify possible directions for further research to improve the model, we perform an error analysis. For a manual analysis of a subset of predictions, we randomly choose 100 mislabeled instances. We identify 10 different categories of mispredictions, which we quantify for our analysis shown in Table 9. The 10 categories are defined as follows:

1. texts in the images do not completely match the texts provided with the dataset (e.g., Figure 8);
2. sarcastic texts, i.e. texts suggesting the opposite emotion to the actual (e.g., Figure 9);
3. sarcastic images, i.e. images suggesting the opposite emotion to the actual (e.g., Figure 10);
4. images that combine multiple concepts, which are difficult for the model to link (e.g., Figure 11);
5. examples requiring real-world knowledge (e.g., Figure 12);
6. inconsistency of emotions expressed in the image and in the text (e.g., Figure 13);
7. our manual label differs from the one in the dataset (e.g., Figure 14);
8. text-only memes (e.g., Figure 15);
9. memes conveying multiple emotions (e.g., Figure 16);
10. the image is abstract (e.g., Figure 17).

| errors | (1) | (2) | (3) | (4) | (5) | (6) | (7) | (8) | (9) | (10) |
| --- | --- | --- | --- | --- | --- | --- | --- | --- | --- | --- |
| % of total | 5% | 38% | 26% | 10% | 50% | 11% | 8% | 4% | 19% | 2% |

**Table 9.** The result of error analysis.

From this analysis, we conclude that a number of these cases would require adding more information to the models, such as weighting of the features. For example, we can expect that emotion-bearing words or facial expressions can support an emotion classification system. However, the image or text may play a more important role in some of the cases, hence potentially benefiting from further studying means for weighting these features in each case.

## 6. Conclusions

In this paper, we propose CDEL, a novel hybrid model to make the most of a deep learning model and a clustering algorithm jointly for emotion classification in internet memes. Evaluating on a benchmark meme emotion dataset, Memotion, released as part of the SemEval-2020 Task 8, our model achieves state-of-the-art performance, outperforming all previous models.

The proposed CDEL first extracts face encoding features as they are vital to extract theme features of the meme, and selects the most suitable clustering algorithm based on the facial features. After that, we build the bimodal deep learning classification models combining pre-trained text and image models and select the best one as the optimal emotion classification model. Next, we apply our combination strategy to fuse the features of both the deep learning and the clustering algorithms.

Our work has some limitations and opens up avenues for future research. First, we use the "local optimal" strategy to build the model, which is embodied in the process of selecting clustering algorithms and joint deep learning models. This strategy may cause the deep learning classification model not to obtain a global optimal value. We would need to explore a different method to find the "global optimal". Second, the performance of the CDEL model on the negative class has room for improvement. A possible solution to this may be the use of data augmentation techniques to achieve a better representation of the class. Third, the lack of additional datasets suitable for this task means that we need to restrict our evaluation to the Memotion dataset. In the future, we would like to test our model on other datasets to assess its generalizability. Fourth, adding extra information, such as commonsense, can also be a promising direction to improve the performance on the most difficult memes. Finally, our model can only deal with the static image-with-caption class of meme, but other classes of memes, such as memes without text or memes with dynamic images, are not studied, primarily owing to the lack of suitable datasets to study them. Creation and publication of datasets including other types of memes would be ideal to enable development of models that consider a more diverse set of memes.

## Acknowledgements

This study was supported by the National Natural Science Foundation of China under grant number 72174086; Xiaoyu Guo conducted part of this work while visiting Queen Mary University of London.

## Error Analysis: Examples

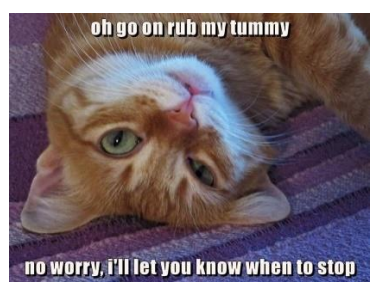

**Figure 8.** Example of error (1): the text provided with the dataset is "oh go on rub my tummy NO no worry", which misses the bottom part of the text in the meme.

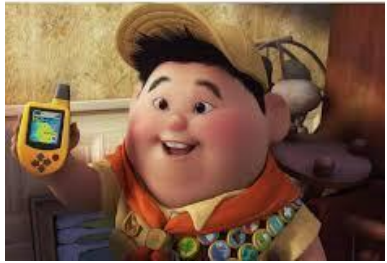

**Figure 9.** Example of error (2): the text contains "launch the nukes", which sarcastically bears a negative emotion, but the actual label is positive.

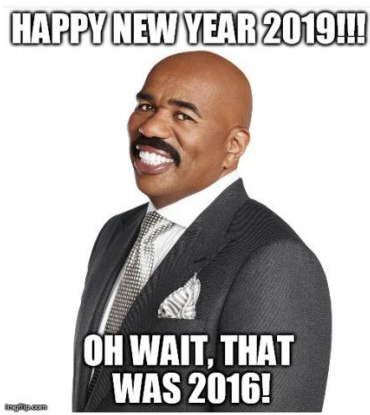

**Figure 10.** Example of error (3): a smiling man in the image suggests a positive emotion, which is however neutral if we consider the entire meme.

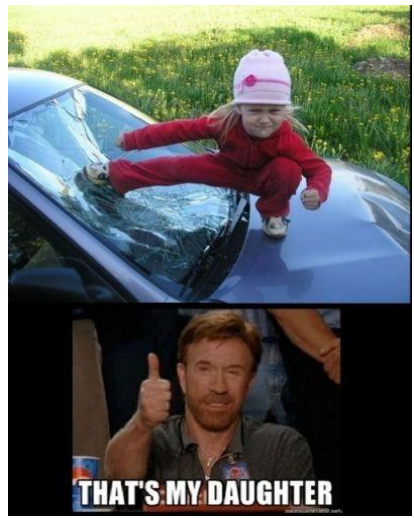

**Figure 11.** Example of error (4): it is difficult to link the positivity of the second image with the act of wrecking a car in the first image.

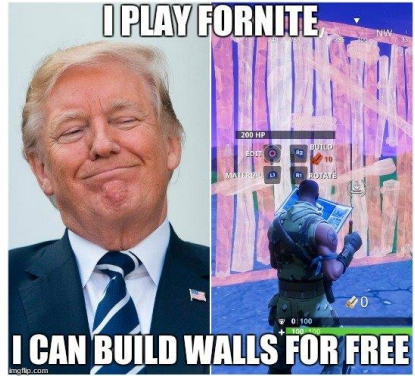

**Figure 12.** Example of error (5): a model needs real-world knowledge to link Donald Trump with the idea of building a wall.

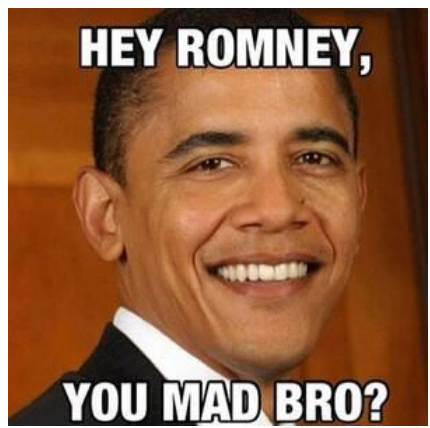

**Figure 13.** Example of error (6): the image bears a positive emotion, whereas the text is negative.

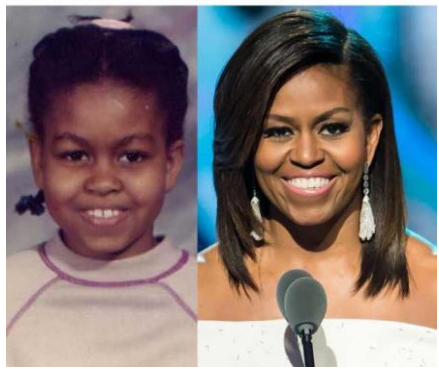

**Figure 14.** Example of error (7): an example that we would label as neutral, whereas the label in the dataset is negative.

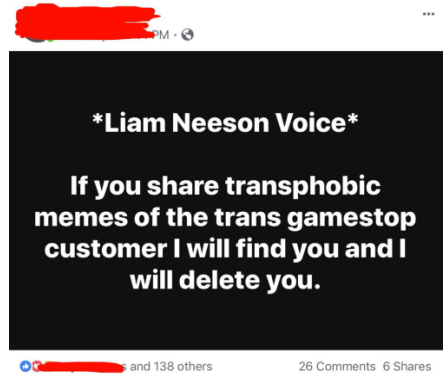

**Figure 15.** Example of error (8): text-only meme.

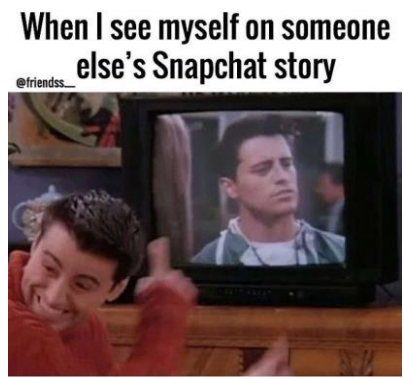

**Figure 16.** Example of error (9): the image mixes two different emotions of the same person.

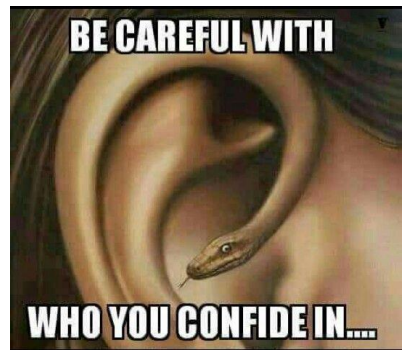

**Figure 17.** Example of error (10): the image is hard to interpret for a model.